\documentclass[final]{cvpr}

\usepackage{times}
\usepackage{epsfig}
\usepackage{graphicx}
\usepackage{amsmath}
\usepackage{amssymb}

\usepackage[pagebackref=true,breaklinks=true,colorlinks,bookmarks=false]{hyperref}

\usepackage[utf8]{inputenc} 
\usepackage[T1]{fontenc}    
\usepackage{url}            
\usepackage{booktabs}       
\usepackage{amsfonts}       
\usepackage{nicefrac}       
\usepackage{microtype}      
\usepackage[font=small,labelfont=bf]{caption}
\usepackage{subcaption}

\usepackage{xcolor}
\usepackage{textgreek}
\usepackage{multirow}
\usepackage{array}
\usepackage[ruled,vlined]{algorithm2e}
\usepackage{tablefootnote}
\usepackage{enumitem}
\usepackage{comment}
\usepackage{diagbox}

\newcommand\Tstrut{\rule{0pt}{2.6ex}}         
\newcommand\Bstrut{\rule[-0.9ex]{0pt}{0pt}}   
 
\newcommand{\NAS}{Neural Architecture-Recipe Search}
\newcommand{\Nas}{neural architecture-recipe search}
\newcommand{\nas}{NARS}
\newcommand{\net}{FBNetV3}
\newcommand{\autotrain}{recipe-only}
\newcommand{\AutoTrain}{Recipe-only}
\newcommand{\predictor}{predictor}
\newcommand{\Predictor}{Predictor}
\newcommand{\NAF}{\Predictor}
\newcommand{\naf}{predictor}
\newcommand{\cio}{constrained iterative optimization}
\newcommand{\CIO}{Constrained Iterative Optimization}
\newcommand{\Cio}{Constrained iterative optimization}
\newcommand{\pbes}{predictor-based evolutionary search}
\newcommand{\Pbes}{Predictor-based evolutionary search}
\newcommand{\PBES}{Predictor-Based Evolutionary Search}

\def\cvprPaperID{4956} 
\def\confYear{CVPR 2021}

\begin{document}

\title{\net: Joint Architecture-Recipe Search using \NAF~Pretraining}


\author{
  \large{Xiaoliang Dai$^{1*}$, Alvin Wan$^{2}$\thanks{Equal contribution},~ Peizhao Zhang$^{1*}$, Bichen Wu$^{1}$, Zijian He$^{1}$, Zhen Wei$^{3}$},  \\\large{Kan Chen$^{1}$, Yuandong Tian$^{1}$, Matthew Yu$^{1}$, Peter Vajda$^{1}$, and Joseph E. Gonzalez$^{2}$} \\
  \normalsize{$^{1}$Facebook Inc., $^{2}$UC Berkeley, $^{3}$UNC Chapel Hill}\vspace{-0.03in}\\
\footnotesize\{\texttt{xiaoliangdai,stzpz,wbc,zijian,kanchen18,yuandong,mattcyu,vajdap\}@fb.com} \vspace{-0.07in}\\
\footnotesize{\texttt{
\{alvinwan,jegonzal\}@berkeley.edu, zhenni@cs.unc.edu}}
}

\maketitle

\begin{abstract}
Neural Architecture Search (NAS) yields state-of-the-art neural networks that outperform their best manually-designed counterparts. However, previous NAS methods search for architectures under one set of training hyperparameters (i.e., a training recipe),
overlooking superior architecture-recipe combinations. To address this, we present \NAS~(\nas)~to search both (a) architectures and (b) their corresponding training recipes, simultaneously. \nas~utilizes an accuracy \naf~that scores architecture \textbf{and} training recipes jointly, guiding both sample selection and ranking.
Furthermore, to compensate for the enlarged search space, we leverage ``free'' architecture statistics (e.g., FLOP count) to pretrain the \naf,
significantly improving its sample efficiency and prediction reliability.
After training the \naf~via constrained iterative optimization, we run fast evolutionary searches in just CPU minutes to generate architecture-recipe pairs for a variety of resource constraints, called \net. \net~makes up a family of state-of-the-art compact neural networks that outperform both automatically and manually-designed competitors. For example, \net~matches both EfficientNet and ResNeSt accuracy on ImageNet with up to 2.0$\times$ and 7.1$\times$ fewer FLOPs, respectively.
Furthermore, \net~yields significant performance gains for downstream object detection tasks, improving mAP despite 18\% fewer FLOPs and 34\% fewer parameters than EfficientNet-based equivalents.
\end{abstract}

\section{Introduction}

Designing efficient computer vision models is a challenging but important problem: A myriad of applications from autonomous vehicles to augmented reality require compact models that must be highly accurate -- even under constraints on power, computation, memory, and latency. The number of possible constraint and architecture combinations is combinatorially large, making manual design a near impossibility.

In response, recent work employs neural architecture search (NAS) to design state-of-the-art efficient deep neural networks. One category of NAS is differentiable neural architecture search (DNAS). These path-finding algorithms are efficient, often completing a search in the time it takes to train one network. However, DNAS cannot search for non-architecture hyperparameters, which are crucial to the model's performance. Furthermore, supernet-based NAS methods suffer from a limited search space, as the entire supergraph must fit into memory to avoid slow convergence \cite{proxylessnas} or paging. Other methods include reinforcement learning (RL)~\cite{mnasnet}, and evolutionary algorithms (ENAS)~\cite{evolution}. However, these methods share several drawbacks:

\begin{figure}[t]
    \centering
    \includegraphics[width=82mm]{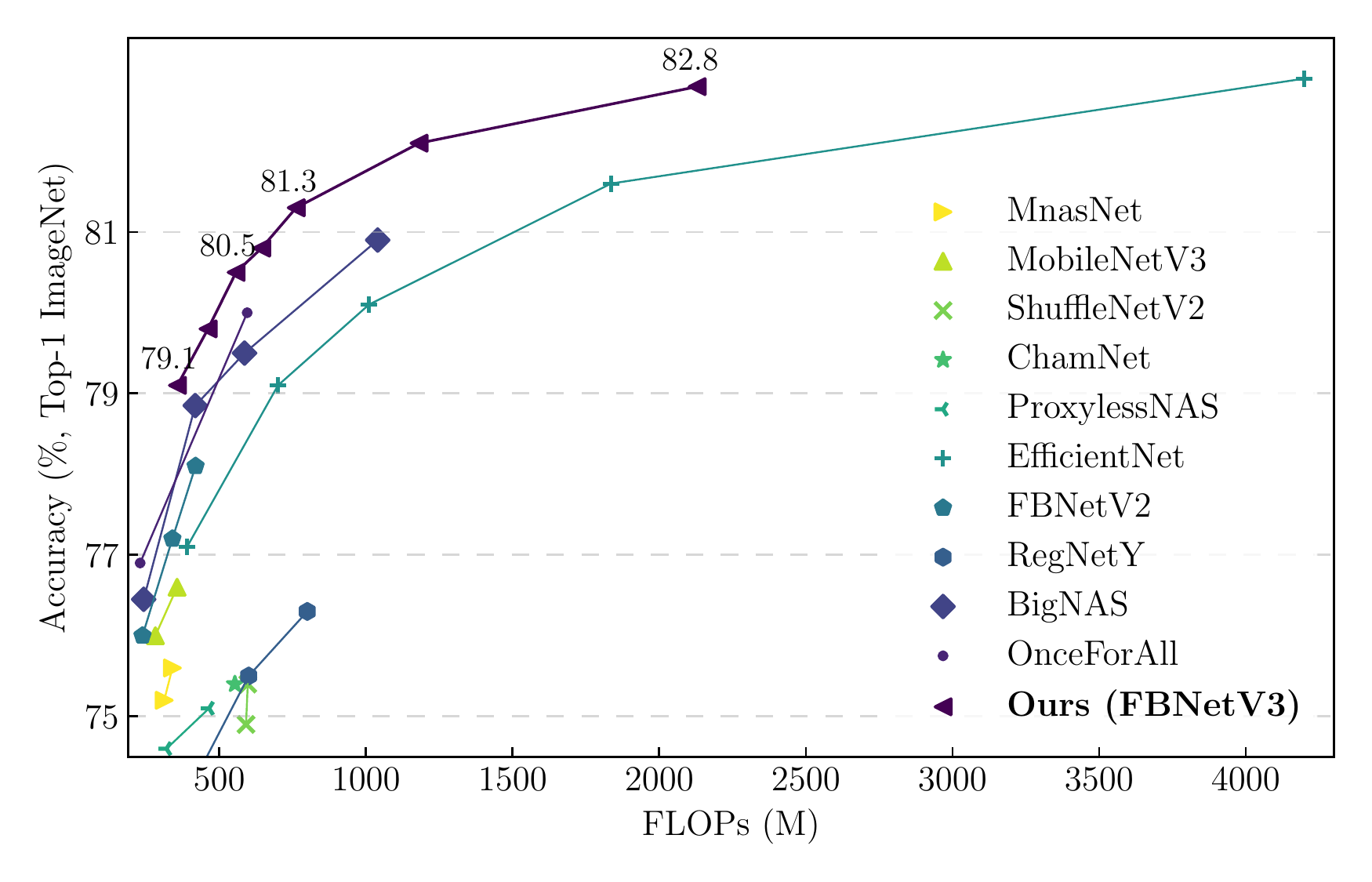}
    \caption{ImageNet accuracy vs. model FLOPs comparison of \net~with other efficient convolutional neural networks. \net~achieves 80.8\% (82.8\%) top-1 accuracy with 557M (2.1G) FLOPs,  setting a new SOTA for accuracy-efficiency trade-offs.}
    \label{fig:imagenet_res}
\end{figure}

\begin{table}
\centering
\footnotesize
\begin{tabular}{cccc}
\toprule
\diagbox[width=8em]{Model}{Training} & Recipe-1 & Recipe-2 \\
\midrule
ResNet18 (1.4x width) & \textbf{70.8\%} & 73.3\% \\
ResNet18 (2x depth) & 70.7\% & \textbf{73.8\%} \\
\bottomrule
\end{tabular}
\caption{Different training recipe could switch the ranking of architectures.  ResNet18 1.4x width and 2x depth refer to ResNet18 with 1.4 width and 2.0 depth scaling factor, respectively.  Training recipe details can be found in Appendix \ref{sec:train-recipe-details}.}
\label{tab:model_recipe} 
\end{table}

\begin{enumerate}[leftmargin=6mm]
    \item \textbf{Ignore training hyperparameters}: NAS, true to its name, searches only for architectures but not the associated training hyperparameters (i.e., ``training recipe''). This ignores the fact that different training recipes may drastically change the success or failure of an architecture, or even switch architecture rankings (Table~\ref{tab:model_recipe}).
    \item \textbf{Support only one-time use}: Many conventional NAS approaches produce one model for a specific set of resource constraints. This means that deploying to a line of products, each with different resource constraints, requires rerunning NAS once for each resource setting. Alternatively, model designers may search for one model and scale it suboptimally, using manual heuristics, to fit new resource constraints.
    \item \textbf{Prohibitively large search space to search:} Naïvely including training recipes in the search space is either impossible (DNAS, supernet-based NAS) or prohibitively expensive, as architecture-only accuracy predictors are already computationally expensive to train (RL, ENAS). 
\end{enumerate}

To overcome these challenges, we propose \NAS~(\nas)~to address the above limitations.  Our insight is three-fold: (1) To support re-use of NAS results for multiple resource constraints, we train an accuracy \naf, then use the \predictor~to find architecture-recipe pairs for new resource constraints in just CPU minutes. (2) To avoid the pitfalls of architecture-only or recipe-only searches, this \naf~scores both training recipes and architectures simultaneously. (3) To avoid prohibitive growth in predictor training time, we pretrain the \naf~on proxy datasets to predict architecture statistics (\eg, FLOPs, \#Parameters) from architecture representations.  After sequentially performing predictor pretraining, constrained iterative optimization, and predictor-based evolutionary search, \nas~produces generalizable training recipes and compact models that attain state-of-the-art performance on ImageNet, outperforming all the existing manually designed or automatically searched neural networks. We summarize our contributions below:

\begin{figure}[t]
    \centering
    \includegraphics[width=60mm]{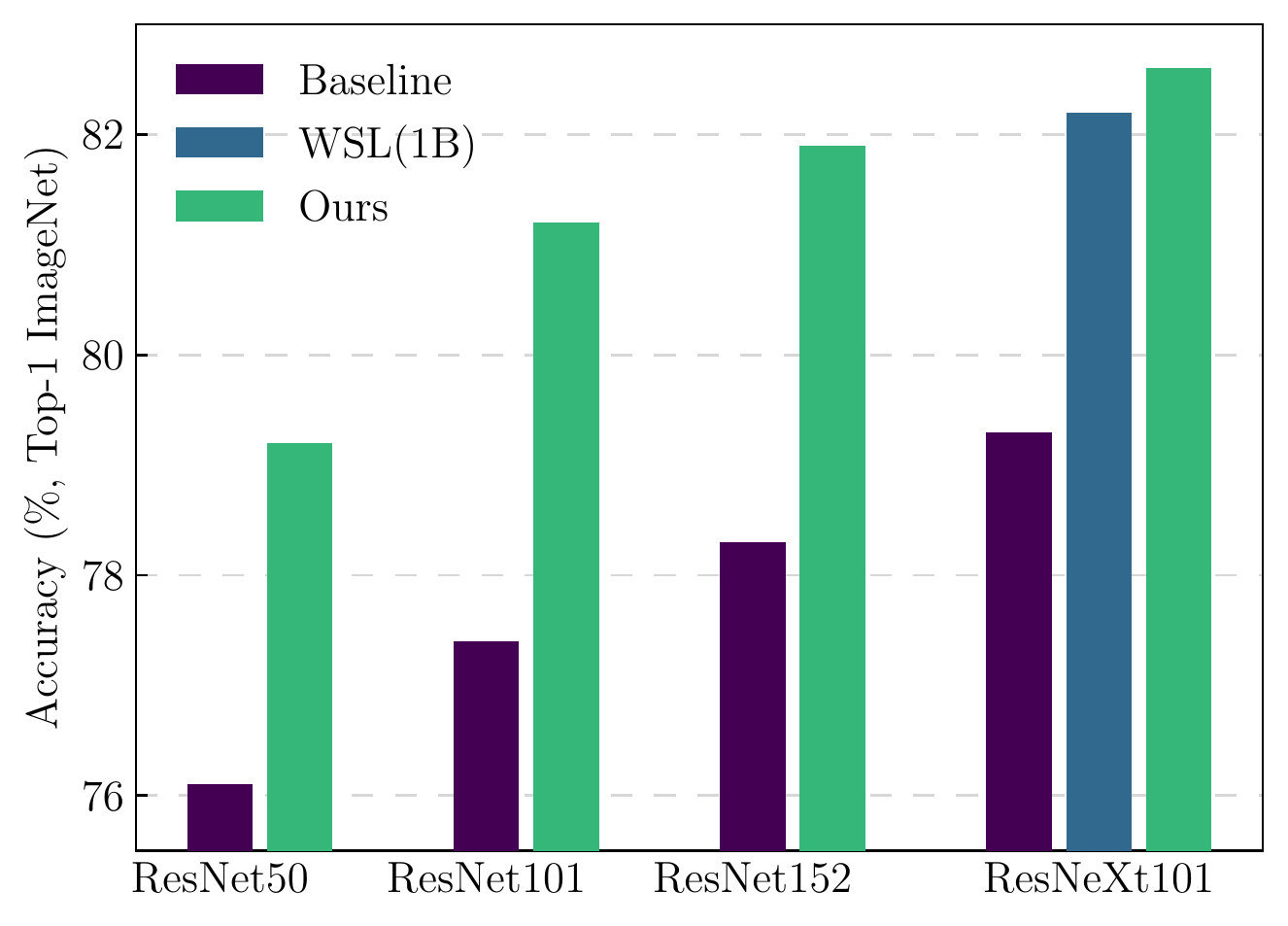}
    \footnotesize
    \caption{Accuracy improvement on existing architectures with the searched training recipe.  WSL refers to the weakly supervised learning model using 1B additional images~\cite{wsl}.}
    \label{fig:training_res}
\end{figure}

\begin{enumerate}[leftmargin=6mm]
    \item \textbf{\NAS}: We propose a predictor that jointly scores both training recipes and architectures, the first joint search, over \textit{both} training recipes and architectures, at scale to our knowledge.
    \item \textbf{Predictor pretraining}: To enable efficient search over this larger space, we furthermore present a pretraining technique, significantly improving the accuracy \predictor's sample efficiency.
    \item \textbf{Multi-use \predictor}: 
    Our \predictor~can be used in fast evolutionary searches to quickly generate models for a wide variety of resource budgets in just CPU minutes.
    \item \textbf{State-of-the-art ImageNet accuracy} per FLOP for the searched \net~models. For example, our \net~matches EfficientNet accuracy with as low as 49.3\% fewer FLOPs, as shown in Fig.~\ref{fig:imagenet_res}.
    \item \textbf{Generalizable training recipe}: \nas's recipe-only search achieves significant accuracy gains across various neural networks, as illustrated in Fig.~\ref{fig:training_res}. Our ResNeXt101-32x8d achieves 82.6\% top-1 accuracy; this even outperforms its weakly-supervised counterpart trained on 1B extra images~\cite{wsl}. 
\end{enumerate}

\section{Related work}

Work on compact neural networks began with manual design, which can be divided into architectural and non-architectural modifications. 

\textbf{Manual architecture design}: Most early work compresses \textit{existing} architectures. One method is pruning~\cite{deepcompression, nest, dreaming, onceforall}, where either layers or channels are removed according to certain heuristics. However, pruning either considers only one architecture~\cite{netprune} or can only sequentially search smaller and smaller architectures~\cite{netadapt}. This limits the search space. Other work designs \textit{new} architectures from the ground up, using new operations that are cost-friendly. This includes convolutional variants like the depthwise convolutions in MobileNet; inverted residual blocks in MobileNetV2; activations such as hswish in MobileNetV3~\cite{mobilenet, mobilenetv2, mobilenetv3}; and operations like shift~\cite{shift} and shuffle~\cite{shufflenetv2}. Although many of these are still used in state-of-the-art neural networks, manually-designed \textit{architectures} have been superseded by automatically-searched counterparts.

\textbf{Non-architectural modifications}: A number of network compression techniques include low-bit quantization~\cite{deepcompression} to as few as two~\cite{tenary} or even one bit~\cite{binary}. Other work downsamples input non-uniformly~\cite{squeezeseg, squeezesegv3, efficientseg} to reduce computational cost. These methods can be combined with architecture improvements for roughly additive reduction in latency.  Other non-architecture modifications involve hyperparameter tuning, including tuning libraries from the pre-deep-learning era \cite{hyperopt}. Several deep-learning-specific tuning libraries are also widely used \cite{raytune}. A newer category of approaches automatically searches for the optimal combination of data augmentation strategies. These methods use policy search \cite{autoaugment}, population-based training~\cite{population}, Bayesian-based augmentation \cite{bayesianaug}, or Bayesian optimization~\cite{kandasamy2018neural}.

\textbf{Automatic architecture search}: NAS automates neural network design for state-of-the-art performance. Several of the most common techniques for NAS include reinforcement learning~\cite{NASRL, mnasnet}, evolutionary algorithms~\cite{evolution,real2019regularized,yang2020cars}, and DNAS~\cite{darts, fbnet, fbnetv2, one-shot, snas}. DNAS trains quickly with few computational resources but is limited by search space size due to memory constraints. Several works seek to address this issue, by training only subsets at a time~\cite{proxylessnas} or by introducing approximations \cite{fbnetv2}. However, its flexibility is still less than that of rival reinforcement learning methods and evolutionary algorithms. In turn, these prior works search for only the model architecture~\cite{progressive,neuralpredictor,wang2020neural,shi2019efficient,onceforall} or perform \Nas~searches on small-scale datasets (e.g., CIFAR) \cite{baker2017accelerating,zela2018towards}. By contrast, our \nas~ jointly searches both architectures and training recipes on ImageNet. To compensate for the larger search space, we (a) introduce a \predictor~pretraining technique to improve the \predictor's~rate of convergence and (b) employ \pbes~to design architecture-recipe pairs in just CPU minutes, for any resource constraint setting--outperforming the \predictor's highest-ranked candidate before evolutionary search significantly. We also note prior work that generates a family of models with negligible or no cost after one search~\cite{one-shot,yang2020cars,lu2020nsganetv2}.


\section{Method}

Our goal is to find the most accurate architecture and training recipe combination, to avoid overlooking architecture-recipe pairs as prior methods have. However, the search space is typically combinatorially large, making exhaustive evaluation an impossibility. To address this, we train an accuracy predictor that accepts architecture and training recipe representations (Sec \ref{sec:naf}). To do so, we employ a three-stage pipeline (Algorithm~\ref{tab:algorithm}): (1)
Pretrain the \predictor~using architecture statistics, significantly improving its accuracy and sample efficiency (Sec \ref{sec:pretrain}). (2) Train the \predictor~using \cio~(Sec \ref{sec:coarse-grained}). (3) For each set of resource constraints, run \pbes~in just CPU minutes to produce high-accuracy architecture-recipe pairs (Sec \ref{sec:fine-grained}). 
 
 \begin{algorithm}[ht]
\footnotesize
\SetAlgoLined
\textbf{Input:}\\  
$\Omega$: the designed search space\; 
$n$: size of candidate pool $\Lambda$ in constrained iterative optimization;\\
$m$: the number of DNN candidates ($\mathcal{X}$) to train in each iteration\;
$T$: the number of batches for \cio; \\
\textbf{Stage 1: Pretrain \NAF} \\
Generate a pool $\Lambda$ with $n$ samples with QMC sampling from the search space $\Omega$;\\
Pretrain accuracy predictor $u$ with architecture statistics;\\
\textbf{Stage 2: Train Predictor (\CIO):} \\
Initialize $\mathcal D_{0}$ as $\emptyset$; \\
\For {$t = 1, 2, ..., T$}{
    Find a batch of the most promising DNN candidates $\mathcal{X}\subset \Lambda$ based on predicted scores, $u(x)$;

    Evaluate all $x\in \mathcal{X}$ by training in parallel; 

    \textbf{if} $t = 1$: Determine early stopping criteria;
    
    Update the dataset: $\mathcal D_{t} = \mathcal D_{t-1} \cup \{(x_1, acc(x_1)), (x_2, acc(x_2)), ...\}$\;

    Retrain the accuracy \predictor~$u$ on $\mathcal D_{t}$\;
}
\textbf{Stage 3: Use Predictor (\PBES)}\\
Initialize $\mathcal{D^{\star}}$ with $p$ best-performing samples in $\mathcal{D}_{T}$ and $q$ randomly generated samples paired with scores predicted by $u$; \\
Initialize $s^{\star}$ with the best score in $\mathcal{D^{\star}}$; set $s^{\star}_0 = 0$; set $\epsilon = 10^{-6}$\;
 \While{$(s^{\star} - s^{\star}_0) >\epsilon$}{
    \For {$x \in \mathcal{D^{\star}}$}{
        Generate a set of children $\mathcal{C}
        \subset \Omega$ subject to resource constraints, by the adaptive genetic algorithm~\cite{chamnet};
    }
    Augment $\mathcal{D^{\star}}$ with $\mathcal{C}$ paired with scores predicted by $u$\;
    
    Select top $K$ candidates from the augmented set to update $\mathcal{D^{\star}}$\;
    
    Update the previous best ranking score by $s_0^{\star} = s^{\star}$\;
    Update the current best ranking score $s^{\star}$ by the best predicted score in $\mathcal{D^{\star}}$.
 }
 \KwResult{$\mathcal{D^{\star}}$, i.e., all the top $K$ best samples with their predicted scores.}
 
 \caption{Three-stage Constraint-aware \NAS}
 \label{tab:algorithm}
 \end{algorithm}

\subsection{\NAF}
\label{sec:naf}

Our \predictor~aims to predict accuracy given representations of an architecture and a training recipe. 
The architecture and training recipe are encoded using one-hot categorical variables (e.g., for block types) and min-max normalized continuous values (e.g., for channel counts). See the full search space in Table \ref{tab:search_space}.

\begin{figure}[t]
 \centering
    \includegraphics[trim=0 0 0 10, clip, width=82mm]{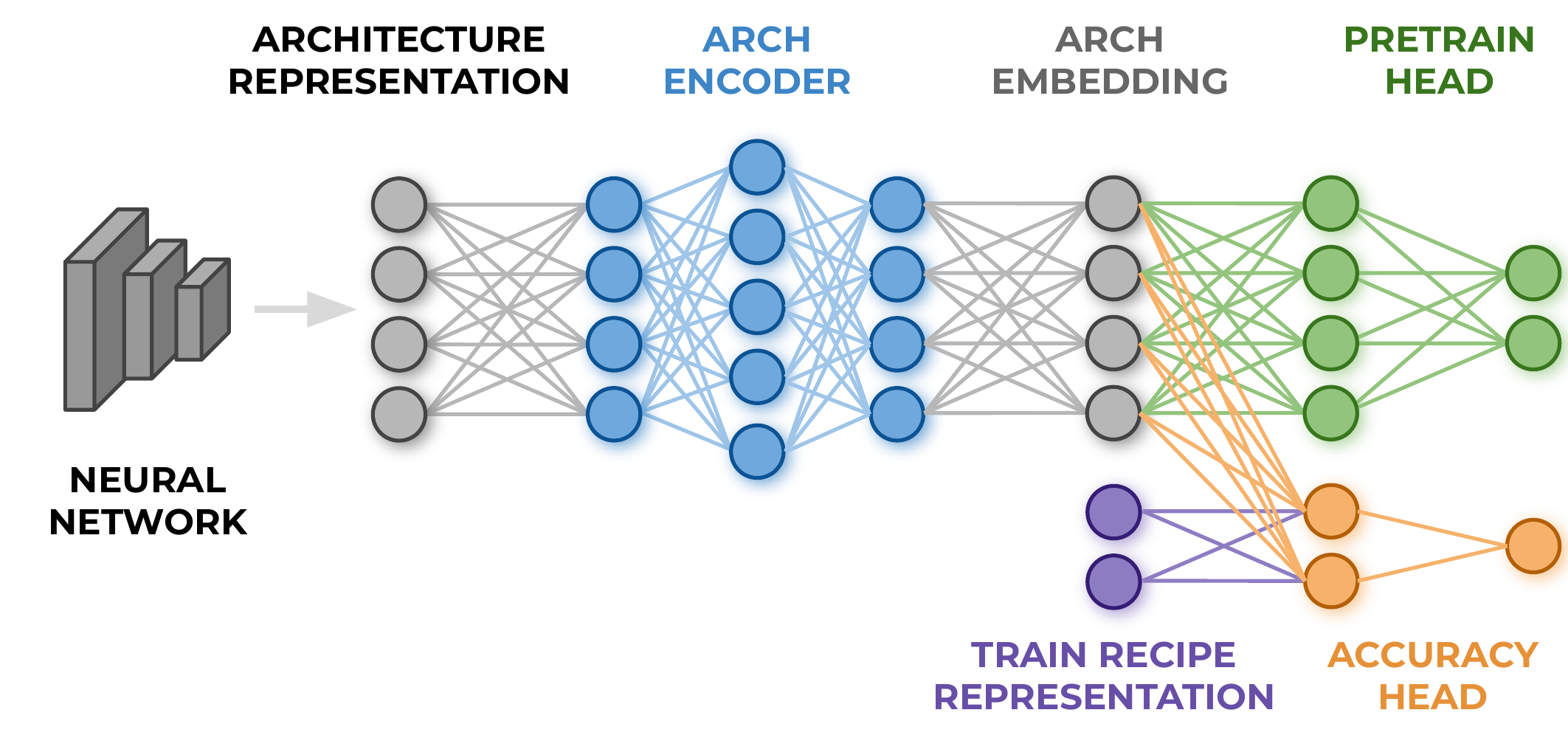}
    \footnotesize
    \caption{\protect\centering Pretrain to predict architecture statistics (top). Train to predict accuracy from architecture-recipe pairs (bottom)}\label{fig:predictor}%
\end{figure}

The \predictor~architecture is a multi-layer perceptron (Fig.~\ref{fig:predictor}) consisting of several fully-connected layers and two heads: (1) An auxiliary ``proxy'' head, used for pretraining the encoder, predicts architecture statistics (e.g., FLOPs and \#Parameters) from architecture representations; and (2) the accuracy head, fine-tuned in \cio~(Sec \ref{sec:coarse-grained}), predicts accuracy from joint representations of the architecture and training recipe.

\subsection{Stage 1: \Predictor~pretraining} 
\label{sec:pretrain}

Training an accuracy predictor can be computationally expensive, as each training label is ostensibly a fully-trained architecture under a specific training recipe. To alleviate this, our insight is to first pretrain on a proxy task. The pretraining step can help the predictor to form a good internal representation of the inputs, therefore reducing the number of accuracy-architecture-recipe samples needed. This can significantly mitigate the search cost required.

\begin{figure*}[t]
\centering
\begin{tabular}{ccccccc}
\centering
    \includegraphics[width=28mm]{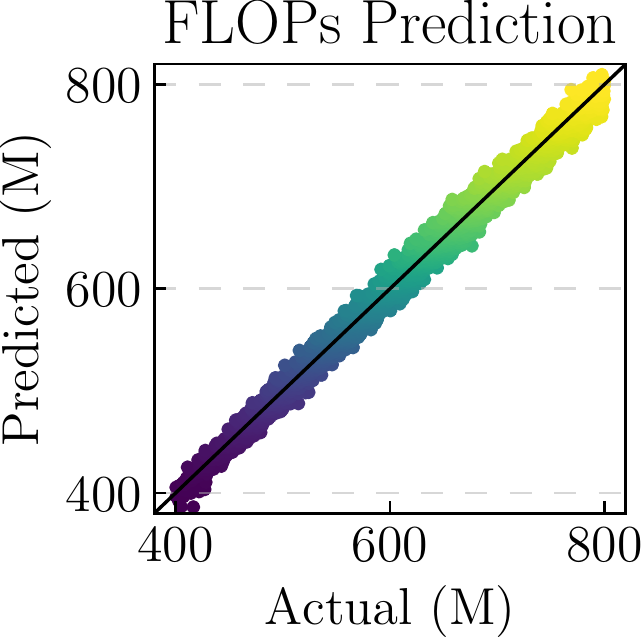} &
    \includegraphics[width=28mm]{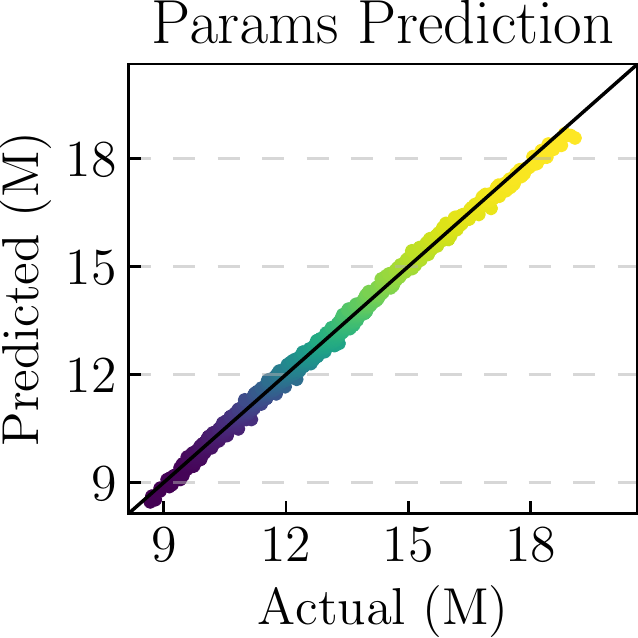} &
    \includegraphics[width=29mm]{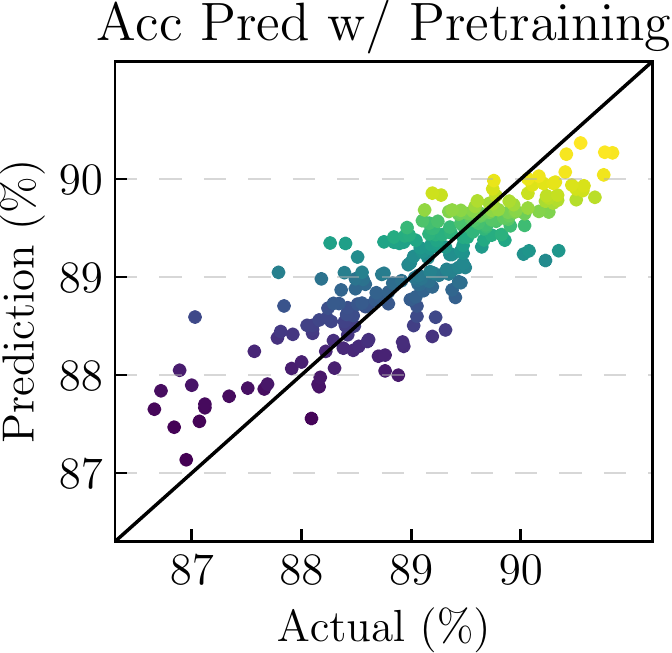}&
    \includegraphics[width=29mm]{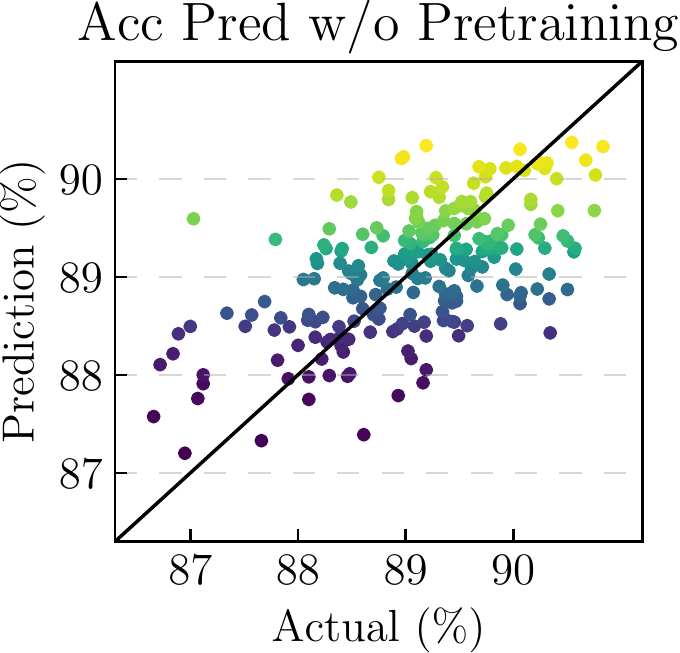}&
    \includegraphics[width=27mm]{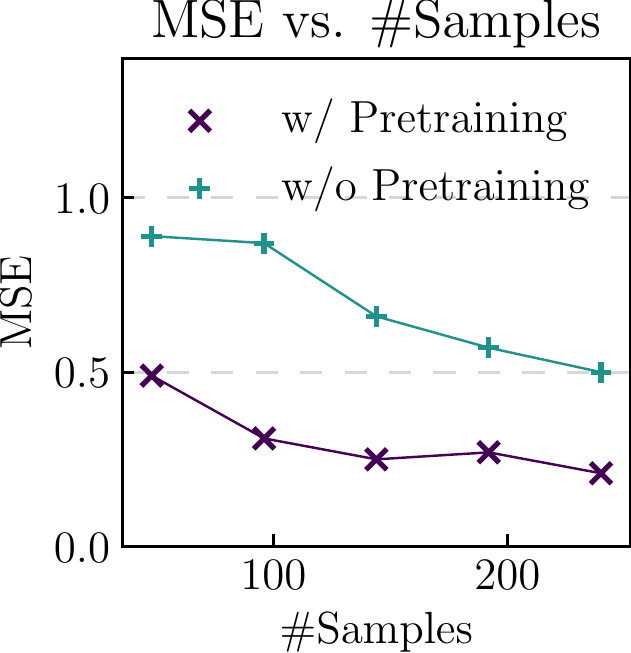}&\\
    (a) & (b) & (c) & (d) & (e) \\
\end{tabular}
\caption{\protect\centering (a) and (b): \Predictor's performance on the proxy metrics, (c) and (d): \Predictor's performance on accuracy with and without pretraining, (e): \Predictor's MSE vs. number of samples with and without pretraining.}
\label{fig:predictor-graphs}%
\end{figure*}

To construct a proxy task for pretraining, we can use ``free'' source of labels for architectures: namely, architecture statistics like FLOPs and numbers of parameters. After this pretraining step, we transfer the pretrained embedding layer to initialize the accuracy \predictor~(Fig.~\ref{fig:predictor}). This leads to significant improvements in the final predictor's sample efficiency and prediction reliability.  For example, to reach the same prediction mean square error (MSE), the pretrained predictor only requires 5$\times$ less samples than its counterpart without pretraining, as shown in Fig.~\ref{fig:predictor-graphs}(e). As a result, predictor pretraining reduces the overall search cost substantially.

\subsection{Stage 2: Training \predictor}
\label{sec:coarse-grained}

In this step, we train the \predictor~and generate a set of high-promise candidates. As mentioned prior, our goal is to find the most accurate architecture and training recipe combination under given resource constraints. We thus formulate the architecture search as a constrained optimization problem:
\begin{equation}\label{eq:constraints}
\begin{aligned}
  \underset{(A, h) \in \Omega}{\operatorname{max}} \;\; acc(A, h), \; 
  \operatorname{s.t.}\; g_i(A) \leqslant C_i,\;\; i = 1, ..., \gamma \\
\end{aligned}
\end{equation}
where $A$, $h$, and $\Omega$ refer to the neural network architecture, training recipe, and designed search space, respectively. $acc$ maps the architecture and training recipe to accuracy.  $g_i(A)$ and $\gamma$ refer to the formula and count of resource constraints, such as computational cost, storage cost, and run-time latency.

\textbf{\Cio}: We first use Quasi Monte-Carlo (QMC) \cite{qmcmc} sampling to generate a sample pool of architecture-recipe pairs from the search space. Then, we train the \naf~iteratively: We (a) shrink the candidate space by selecting a subset of favorable candidates based on predicted accuracy, (b) train and evaluate the candidates using an early-stopping heuristic, and (c) fine-tune the \predictor~with the Huber loss. This iterative shrinking of the candidate space avoids unnecessary evaluations and improves exploration efficiency.

\begin{itemize}[leftmargin=6mm]

\begin{figure}[t]
\centering
\includegraphics[width=60mm]{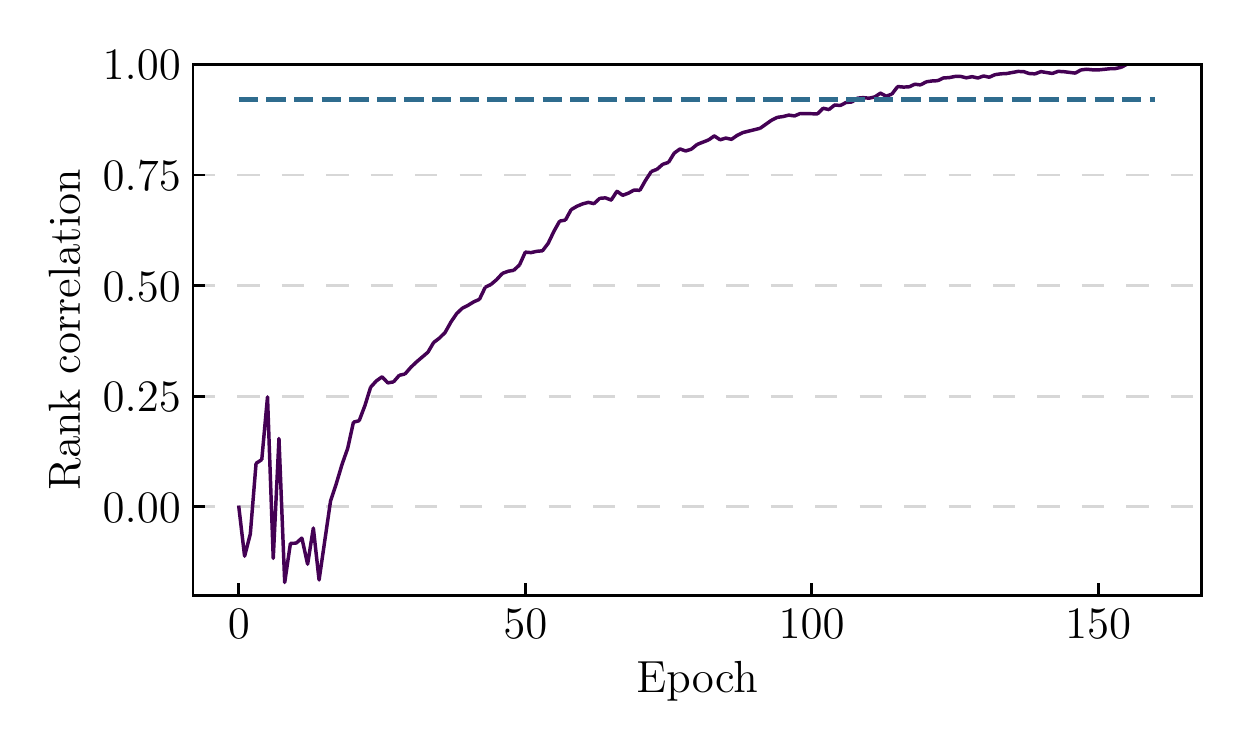}
    \footnotesize
    \caption{\protect\centering Rank correlation vs. epochs. Correlation threshold (cyan) is 0.92. 
    }
    \label{fig:early_stop}%
\end{figure}

\item \textbf{Training candidates with early-stopping}. We introduce an early stopping mechanism to cut down on the computational cost of evaluating candidates. Specifically, we (a) rank samples by both early-stopping and final accuracy after the first iteration of \cio, (b) compute the rank correlation, and (c) find the epoch $e$ where correlation exceeds a particular threshold (e.g., 0.92), as shown in Fig.~\ref{fig:early_stop}. 

For all remaining candidates, we train ($A, h$) only for $e$ epochs to approximate $acc(A, h)$. This allows us to use much fewer training iterations to evaluate each queried sample.

\item \textbf{Training the \predictor~with Huber loss}. After obtaining the pretrained architecture embedding, we first train the predictor for 50 epochs with the embedding layer frozen. Then, we train the entire model with reduced learning rate for another 50 epochs. We adopt the Huber loss to train the accuracy \predictor, i.e., $\mathcal{L} = 0.5(y - \hat{y})^{2}$ if $|y - \hat{y}| < 1$ else $|y - \hat{y}| - 0.5$, where $y$ and $\hat{y}$ are the prediction and ground truth label, respectively. This prevents the model from being dominated by outliers, which shows can confound the predictor~\cite{neuralpredictor}. 
\end{itemize}

\subsection{Stage 3: Using \predictor}
\label{sec:fine-grained}

The third stage of the proposed method is an iterative process based on adaptive genetic algorithms~\cite{adaptiveGA}.
The best-performing architecture-recipe pairs from the second stage are inherited as part of the first generation candidates.  
In each iteration, we introduce mutations to the candidates and generate a set of children $\mathcal{C} \subset \Omega$ subject to given constraints. 
We evaluate the score for each child with the pretrained accuracy \predictor~$u$, and select top $K$ highest-scoring candidates for the next generation.
We compute the gain of the highest score after each iteration, and terminate the loop when the improvement saturates.  
Finally, the \pbes~produces high-accuracy neural network architectures and training recipes.

Note that with the accuracy \predictor, searching for networks to fit different use scenarios only incurs negligible cost.  This is because the accuracy \predictor~can be substantially reused under different resource constraints, while \pbes~takes just CPU minutes.  

\subsection{\Predictor~search space}
Our search space consists of both training recipes and architecture configurations. The search space for training recipes features optimizer type, initial learning rate, weight decay, mixup ratio~\cite{mixup}, drop out ratio, stochastic depth drop ratio~\cite{dropconnect}, and whether or not to use model exponential moving average (EMA)~\cite{kingma2014adam}. Our architecture configuration search space is based on the inverted residual block~\cite{mobilenetv2} and includes input resolution, kernel size, expansion, number of channels per layer, and depth, as detailed in Table~\ref{tab:search_space}.

\begin{table*}
\centering
\footnotesize
\begin{tabular*}{\textwidth}{l @{\extracolsep{\fill}} lllllll}
\toprule
block       & k     &   e   & c     & n     & s  & se & act.\\ \midrule
Conv    & 3     & -     & (16, 24, 2)    & 1     & 2  & -  & hswish\\
MBConv 	 & [3, 5] 	 & 1                & (16, 24, 2) & (1, 4) & 1 & N & hswish\\
MBConv	 & [3, 5] 	 & (4, 7) / (2, 5)  & (20, 32, 4) & (4, 7) & 2 & N & hswish\\
MBConv	 & [3, 5]	 & (4, 7) / (2, 5)	& (24, 48, 4) & (4, 7) & 2 & Y & hswish\\
MBConv	 & [3, 5] 	 & (4, 7)$^{1}$ / (2, 5)$^{2}$  & (56, 84, 4) & (4, 8) & 2 & N & hswish\\
MBConv	 & [3, 5] 	 & (4, 7)$^{1}$ / (2, 5)$^{2}$  & (96, 144, 4)& (6, 10)& 1 & Y & hswish\\
MBConv	 & [3, 5] 	 & (4, 7) 	        & (180, 224, 4)& (5, 9)& 2 & Y & hswish\\
MBConv	 & [3, 5] 	 & 6 	            & (180, 224, 4)& 1     & 1 & Y & hswish\\
MBPool	 & [3, 5] 	 & 6	 & 1984 	 & 1                       &- & - & hswish\\
FC     & -         &  - & 1000        & 1           & - & -  & -\\
\hline \hline
\Tstrut
res & lr($10^{-3}$) & optim & ema & p($10^{-2}$) & d($10^{-1}$) & m($10^{-1}$) & wd($10^{-6}$) \\
\hline
(224, 272, 8) & (20, 30) & [RMSProp, SGD] & [true, false] & (1, 31) & (10, 31) & (0, 41) & (7, 21) \\
 \bottomrule
\end{tabular*}

\caption{
The network architecture configuration and search space in our experiments. MBConv, MBPool, k, e, c, n, s, se, and act. refer to the inverted residual block~\cite{mobilenetv2}, efficient last stage~\cite{mobilenetv3}, kernel size, expansion, \#Channel, \#Layers, stride, squeeze-and-excitation, and activation function, respectively.  res, lr, optim, ema, p, d, m, and wd refer to resolution, initial learning rate, optimizer type, EMA, dropout ratio, stochastic depth drop probability, mixup ratio, and weight decay, respectively. Expansion on the left of the slash is used in the first block in the stage, while that on the right for the rest. Tuples of three values in parentheses represent the lowest value, highest, and steps; two-value tuples imply a step of 1, and tuples in brackets represent all available choices during search. Note that lr is multiplied by 4 if the optim chooses SGD. Architecture parameters with the same superscript share the same values during the search.}
\label{tab:search_space}
\end{table*}

In \autotrain~experiments, we only tune training recipes on a fixed architecture. However, for joint search, we search both training recipes and architectures, within the search space in Table~\ref{tab:search_space}.  Overall, the space contains $10^{17}$ architecture candidates with $10^{7}$ possible training recipes. Exploring such a vast search space for an optimal network architecture and its corresponding training recipe is non-trivial.

\section{Experiments}
In this section, we first validate our search method in a narrowed search space to discover the training recipe for a given network.  Then, we evaluate our search method for joint search over architecture and training recipes. 
We use PyTorch~\cite{pytorch}, and conduct our search on the ImageNet 2012 classification dataset~\cite{imagenet}.  In the search process, we randomly sample 200 classes from the entire dataset to reduce the training time. Then, we randomly withhold 10K images from the 200-class training set as the validation set.

\subsection{Recipe-only search}
\label{sec:ros}

\begin{figure}[h]
    \centering
    \includegraphics[width=65mm]{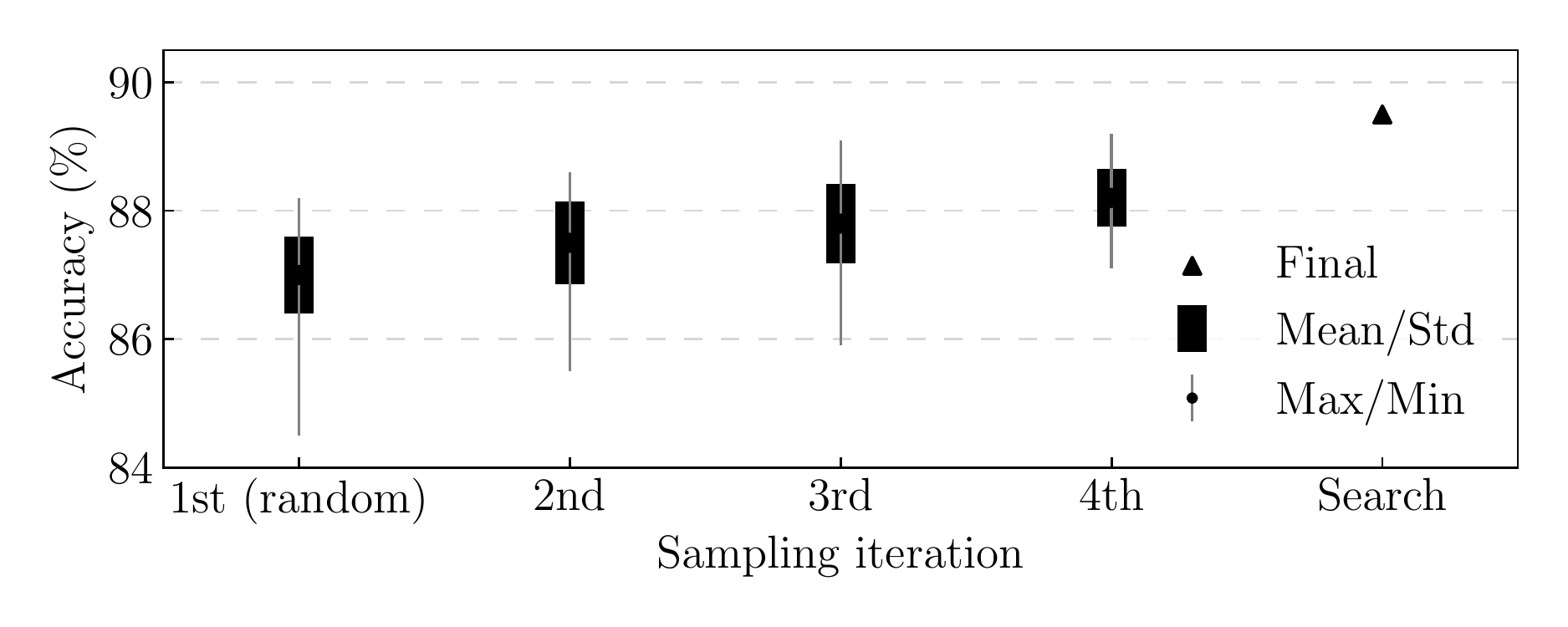}
    \vspace{-2mm}
    \caption{Illustration of the sampling and search process.}
    \label{fig:random-search}
\end{figure}

To establish that even modern NAS-produced architecture’s performance can be further improved with better training recipe, we optimize over training recipes for a fixed architecture. We adopt FBNetV2-L3~\cite{fbnetv2} (Appendix~\ref{sec:fbnetv2_l3}) as our base architecture, which is a DNAS searched architecture that achieves 79.1\% top-1 accuracy with the original training method used in~\cite{fbnetv2}. We set the sample pool size $n = 20K$, batch size $m = 48$ and iteration $T = 4$ in \cio.  We train the sampled candidates for 150 epochs with a learning rate decay factor of 0.963 per epoch during the search, and train the final model with 3$\times$ slower learning rate decay (i.e., 0.9875 per epoch). We show the distribution of samples at each round as well as the final searched result in our experiments in Fig.~\ref{fig:random-search}, where the first-round samples are randomly generated. The searched training recipe (Appendix \ref{sec:detailed-settings}) improves the accuracy of our base architecture by 0.8\%.

We extend the \nas-searched training recipe to other commonly-used neural networks to further validate its generality.  Although the \nas-searched training recipe was tailored to FBNetV2-L3, it generalizes surprisingly well, as shown in Table~\ref{tab:train_extension}. The \nas-searched training recipe leads to substantial accuracy gains of up to 5.7\% on ImageNet. In fact, ResNet50 outperforms the baseline ResNet152 by 0.9\%. ResNeXt101-32x8d even surpasses the weakly supervised learning model, which is trained with 1 billion weakly-labeled images and achieves 82.2\% top-1 accuracy.  Notably, it is possible to achieve even better performance by searching for specific training recipe for each neural network, which would increase the search cost.

\begin{table}
\centering
\footnotesize
    \begin{tabular*}{80mm}{@{}l@{\extracolsep{\fill}}cccc@{}}
    \toprule
    Model  & \multicolumn{3}{c}{Top-1 Accuracy (\%)} \\
    \cmidrule{2-4} & {\scriptsize Original} & {\scriptsize \AutoTrain} & {\scriptsize $\Delta$}\\
    \midrule\Tstrut
    FBNetV2-L3~\cite{fbnetv2} & 79.1 & 79.9 & +0.8\\
    AlexNet~\cite{alexnet} &56.6	&62.3 & +5.7\\
    ResNet34~\cite{resnet} &73.3	&76.3 & +3.0\\ 
    ResNet50~\cite{resnet}   & 76.1  & 79.2 & +3.1 \\
    ResNet101~\cite{resnet}   &77.4	& 81.2 & +3.8 \\
    ResNet152~\cite{resnet}	&78.3	&81.9 & +3.6\\
    DenseNet201~\cite{densenet}	&77.2	&80.2 & +3.0 \\
    ResNeXt101~\cite{resnext} & 79.3 & 82.6 & +3.3 \\
    \bottomrule
    \end{tabular*}
    \captionof{table}{Accuracy improvements with the searched training recipes on existing neural networks. Above, ResNeXt101 refers to the 32x8d variant.} 
    \label{tab:train_extension}
\end{table}

\subsection{\NAS~(\nas)}
\textbf{Search settings} Next, we perform a joint search of architecture and training recipes to discover compact neural networks. Note that based on our observations in Sec.~\ref{sec:ros}, we shrink the search space to always use EMA. Most of the settings are the same as in the \autotrain~search, while we increase the optimization iteration $T = 5$ and set the FLOPs constraint for the sample pool from 400M to 800M.  We pretrain the architecture embedding layer using 80\% of the sample pool which contains 20K samples, and plot the validation on the rest 20\% in~Fig.~\ref{fig:predictor-graphs}.  In the \pbes, we set four different FLOPs constraints: 450M, 550M, 650M, and 750M and discover four models (namely \net-B/C/D/E) with the same accuracy \predictor.  We further scale down and up the minimum and maximum models and generate \net-A and \net-F/G to fit more use scenarios, respectively, with compound scaling proposed in~\cite{efficientnet}.

\textbf{Training setup} For model training, we use a two-step distillation based training process: (1) We first train the largest model (i.e., \net-G) with the searched recipe with ground truth labels.  (2) Then, we train all the models (including \net-G~itself) with distillation, which is a typical training technique adopted in~\cite{onceforall}\cite{bignas}.  Different from the in-place distillation method in~\cite{onceforall}\cite{bignas}, the teacher model here is the ImageNet pretrained \net-G derived from step (1).  The training loss is a sum of two components: Distillation loss scaled by 0.8 and cross entropy loss scaled by 0.2.  During training, we use synchronized batch normalization in distributed training with 8 nodes and 8 GPUs per node. We train the models for 400 epochs with a learning rate decay factor of 0.9875 per epoch after a 5-epoch warmup.  We train the scaled models \net-A and \net-F/G with the searched training recipes for FBNetV3-B and \net-E, respectively, only increasing the stochastic depth drop ratio for \net-F/G to 0.2.  More training details can be found in Appendix~\ref{sec:training_details}.

\newcommand{\scfbnetvThree}{10.7K}
\newcommand{\scfbnetvTwo}{0.6K}
\newcommand{\scfbnet}{0.2K}
\newcommand{\scchamnet}{28K}
\newcommand{\scmobilenetvThree}{>91K}
\newcommand{\scproxyless}{0.2K}
\newcommand{\scefficientnet}{>91K}
\newcommand{\scregnet}{11K} 
\newcommand{\scatomnas}{0.8K} 
\newcommand{\scbignas}{2.3K} 

\newcommand{\ttfbnetvThree}{ademl}
\newcommand{\ttfbnetvTwo}{el}
\newcommand{\ttfbnet}{l}
\newcommand{\ttchamnet}{l}
\newcommand{\ttmobilenetvThree}{el}
\newcommand{\ttproxyless}{ml}
\newcommand{\ttefficientnet}{ademl}
\newcommand{\ttregnet}{-}
\newcommand{\ttatomnas}{el}
\newcommand{\ttbignas}{de}
\newcommand{\ttresnet}{-}
\newcommand{\ttresnest}{ademl}
\newcommand{\ttresnext}{-}
\newcommand{\ttofa}{deml}

\begin{table*}[t]
\centering
\footnotesize
\begin{tabular*}{174mm}{@{}l@{\extracolsep{\fill}}|lcc|cccccc@{}}
\toprule
Model & Search method & Search space & Search cost & FLOPs & Accuracy & Accuracy \\
& & & (GPU/TPU hours) & &  (\%, Top-5) &  (\%, Top-1) \\
\midrule
FBNet~\cite{fbnet} & gradient & arch & \scfbnet &  375M &  - & 74.9 \\
ProxylessNAS~\cite{proxylessnas} & RL/gradient & arch & \scproxyless  & 465M & - & 75.1 \\
ChamNet~\cite{chamnet} & \predictor~& arch & \scchamnet  & 553M & - & 75.4 \\
RegNetY~\cite{radosavovic2020designing} & pop. param.$^{*}$ & arch & \scregnet & 600M & - & 75.5 \\
MobileNetV3-1.25x~\cite{mobilenetv3} & RL/NetAdapt & arch & \scmobilenetvThree  & 356M & - & 76.6 \\
EfficientNetB0~\cite{efficientnet} & RL/scaling & arch & \scefficientnet& 390M  & 93.3 & 77.3 \\
AtomNAS~\cite{atomnas} & gradient & arch & \scatomnas& 363M & - &  77.6 \\
FBNetV2-L2~\cite{fbnetv2} & gradient & arch & \scfbnetvTwo  & 423M & - &  78.1 \\ 
\textbf{\net-A} & \textbf{\nas} & \textbf{arch/recipe} & \scfbnetvThree & \textbf{357M}  & \textbf{94.5} & \textbf{79.1} \\
\midrule
ResNet152~\cite{resnet} & manual & - & -  & 11G & 93.8 & 78.3 \\
EfficientNetB2~\cite{efficientnet} & RL/scaling & arch & \scefficientnet  & 1.0G &  94.9 & 80.3 \\
ResNeXt101-32x8d~\cite{resnext} & manual & - & -  & 7.8G & 94.5 & 79.3 \\
Once-For-All~\cite{onceforall} & gradient & - & -  & 595M & - & 80.0 \\
\textbf{\net-C} & \textbf{\nas} & \textbf{arch/recipe} & \scfbnetvThree &  \textbf{557M} & \textbf{95.1} & \textbf{80.5}   \\
\midrule
BigNASModel-XL~\cite{bignas} & gradient & arch & \scbignas & 1.0G & - & 80.9 \\ 
ResNeSt-50~\cite{resnest} & manual & - & -& 5.4G & - & 81.1 \\
\textbf{\net-E} & \textbf{\nas} & \textbf{arch/recipe} & \scfbnetvThree  & \textbf{762M} & \textbf{95.5} & \textbf{81.3} \\
\midrule
EfficientNetB3~\cite{efficientnet} & RL/scaling & arch & \scefficientnet  & 1.8G & 95.7 & 81.7 \\
ResNeSt-101~\cite{resnest} & manual & - & - & 10.2G & - & 82.3 \\
EfficientNetB4~\cite{efficientnet} & RL/scaling & arch & \scefficientnet  & 4.2G  & 96.4 & 82.9\\
\textbf{\net-G} & \textbf{\nas} & \textbf{arch/recipe} & \scfbnetvThree & \textbf{2.1G}  & \textbf{96.3} & \textbf{82.8}
\\
\bottomrule
\end{tabular*}%
\caption{Comparisons of different compact neural networks. For baselines, we cite statistics on ImageNet from the original papers. Our results are
bolded. *: population parameterization. See \ref{sec:training-tricks-efficientnet} for discussions about the training tricks and additional EfficientNet comparisons.}
\label{tab:imagenet_res}
\end{table*}

\textbf{Searched models} We compare our searched model against other relevant NAS baselines and hand-crafted compact neural networks in Fig.~\ref{fig:imagenet_res}, and list the detailed performance metrics comparison in Table~\ref{tab:imagenet_res}, where we group the models by their top-1 accuracy. Among all the existing efficient models such as EfficientNet~\cite{efficientnet}, MobileNetV3~\cite{mobilenetv3}, ResNeSt~\cite{resnest}, and FBNetV2~\cite{fbnetv2}, our searched model delivers substantial improvements on the accuracy-efficiency trade-off.  For example, on low computation cost regime, \net-A achieves 79.1\% top-1 accuracy with only 357M FLOPs (2.5\% higher accuracy than MobileNetV3-1.25x~\cite{mobilenetv3} with similar FLOPs). On high accuracy regime, \net-E achieves 0.2 higher accuracy with over 7$\times$ fewer FLOPs compared to ResNeSt-50~\cite{resnest}, while \net-G achieves the same level of accuracy as EfficientNetB4~\cite{efficientnet} with 2$\times$ fewer FLOPs. Note that we have further improved the accuracy of FBNetV3 by using larger teacher models for distillation, as shown in Appendix~\ref{sec:further_gain}.

\subsection{Transferability of the searched models}

\begin{figure}[t]
\centering
    \includegraphics[width=0.45\textwidth]{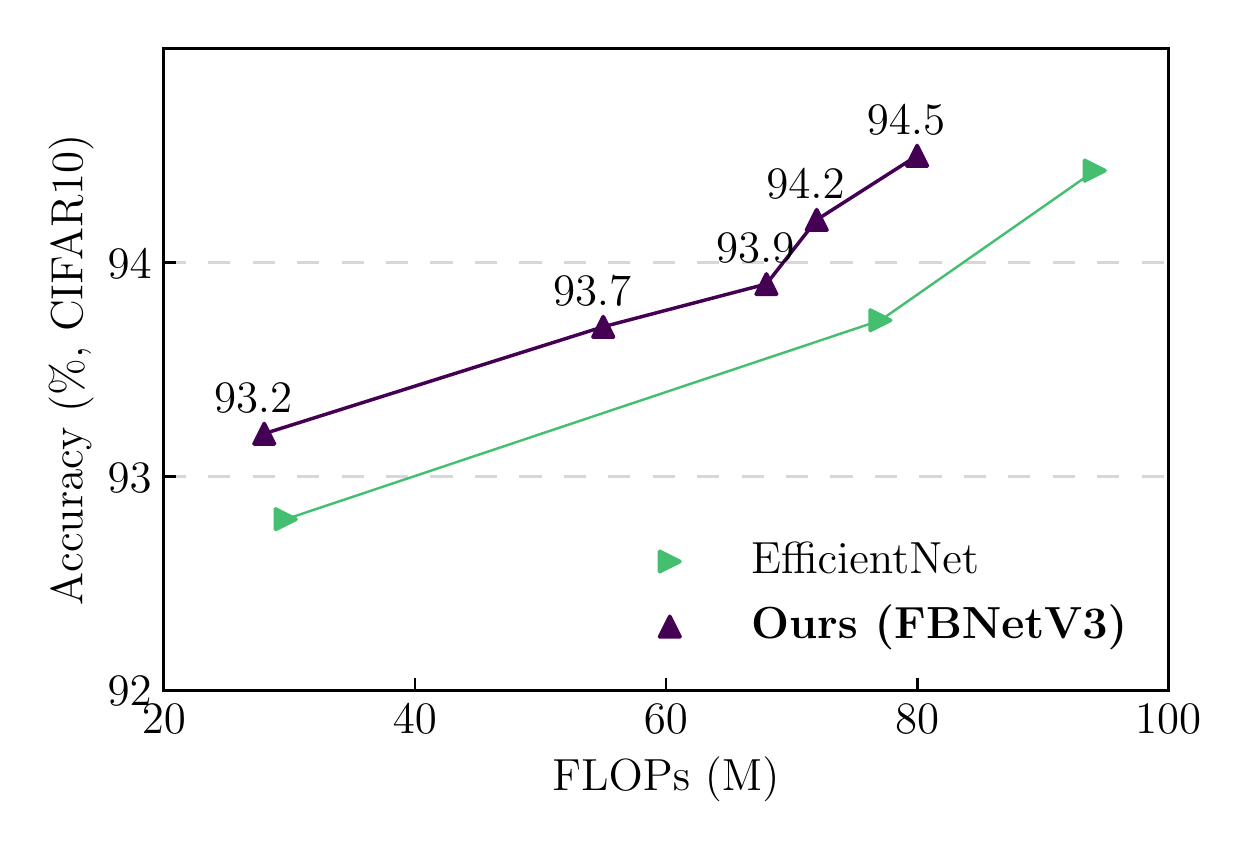}
    \caption{Accuracy vs. FLOPs comparison on the CIFAR-10 dataset.}
    \label{fig:cifar}
\end{figure}

\textbf{Classification on CIFAR-10} We further extend the searched \net~on CIFAR-10 dataset that has 60K images from 10 classes~\cite{cifar} to validate its transferability.  Note that different from~\cite{efficientnet} that scales up the base input resolution to 224$\times$224, we keep the original base input resolution as 32$\times$32, and scale up the input resolutions for larger models based on the scaling ratio.  We also replace the second stride-two block with a stride-one block to fit the low-resolution inputs. We don't include distillation for simplicity. We compared the performance of different models in Fig.~\ref{fig:cifar}. Again, our searched models significantly outperform the EfficientNet baselines.

\begin{table}[h]
\centering
\footnotesize
\begin{tabular}{cccc}
\toprule
Backbone & \#Params (M) & FLOPs (G) & mAP \\
\midrule
EfficientNetB0 & 8.0 & 3.6 & 30.2 \\
\net-A & 5.3 & 2.9  & 30.5 \\
\midrule
EfficeintNetB1& 13.3 & 5.6  & 32.2 \\
\net-E  & 10.6 & 5.3 & 33.0 \\
\bottomrule
\end{tabular}%
\caption{Object detection results of Faster RCNN with different backbones on COCO.}
\label{tab:coco_det} 
\end{table}

\textbf{Detection on COCO}
To further validate the transferability of the searched models on different tasks, we use \net~as a replacement for the backbone feature extractor for Faster R-CNN with the conv4 (C4) backbone and compare with other models on the COCO detection dataset.  We adopt most of the training settings in~\cite{detectron2} with 3$\times$ training iterations, while use synchronized batch normalization, initialize the learning rate at 0.16, switch on EMA, reduce the non-maximum suppression (NMS) to 75, and change to learning rate schedule to Cosine after warming up.  Note that we only transfer the searched architectures and use the same training protocol for all the models.

We show the detailed COCO detection results in Table~\ref{tab:coco_det}. With similar or higher mAP, our \net~reduces the FLOPs and number of parameters by up to 18.3\% and 34.1\%, respectively, compared to EfficientNet backbones.

\section{Ablation study and discussions}
In this section, we revisit the performance improvements obtained from joint search, significance of the \pbes, and the impact and generality of several training techniques.

\textbf{Architecture and training recipe pairing}. Our method yields different training recipes for different models. For example, we observe that smaller models tend to prefer less regularization (e.g., smaller stochastic depth drop ratio and mixup ratio). To illustrate the significance of \Nas, we swap the training recipes searched for \net-B and \net-E, observing a significant accuracy drop for both models, as shown in Table~\ref{tab:swap_training}. This highlights the importance of correct architecture-recipe pairings, emphasizing the downfall of conventional NAS: Ignoring the training recipe and only searching for the network architecture fails to obtain optimal performance.
\begin{table}[h]
\centering
\footnotesize
\begin{tabular}{lcc}
    \toprule
     & \net-B & \net-E \\
     & {\small Train recipe} & {\small Train recipe}\Bstrut \\
    \midrule  
    \Tstrut
    \net-B Arch & \textbf{79.8\%} & 78.5\% \\
    \net-E Arch & 80.8\% & \textbf{81.3\%} \\
    \bottomrule
    \end{tabular}
    \captionof{table}{Accuracy comparison for the searched models with swapped training recipes.}
    \label{tab:swap_training}
\end{table}

\textbf{\Pbes~improvements}.
\Pbes~ yields substantial improvement on top of \cio. To demonstrate this, we compare the best-performing candidates derived from the second search stage with the final searched \net~under the same FLOPs constraints (Table~\ref{tab:fine_grained_search}).  We observe an accuracy drop of up to 0.8\% if the third stage is discarded.  Thus, the third search stage, though requiring only negligible cost (i.e., several CPU minutes), is equally crucial to the final models' performance.
\begin{table}[h]
\centering
\footnotesize
    \begin{tabular}{lccc}
    \toprule
    Model & Evolutionary Search & FLOPs & Accuracy \\
    \midrule
    \net-B & Y & 461M & 79.8\% \\
    \net-B$^{*}$ & N & 448M & 79.0\% \\
    \net-E & Y & 762M & 81.3\% \\
    \net-E$^{*}$ & N & 746M & 80.7\% \\
    \bottomrule
    \end{tabular}
    \captionof{table}{Performance improvement by the \pbes search. *: Models derived from \cio.}
    \label{tab:fine_grained_search}
    \vspace{-1mm}
\end{table}

\vfill\null
\textbf{Impact of distillation and model averaging}
We show the model performance on \net-G in Table~\ref{tab:distill_ema} with different training configurations, where the baseline refers to the vanilla training without EMA or distillation. EMA brings substantially higher accuracy, especially during the middle stage of training.  We hypothesize EMA intrinsically functions as a strong ``ensemble'' mechanism and thus improves single-model accuracy. We additionally observe distillation brings notable performance improvement. This is consistent with the observations in ~\cite{onceforall, bignas}. Note since the teacher is a pretrained \net-G, \net-G~ is self-distilled. The combination of EMA and distillation improves the model's top-1 accuracy from 80.9\% to 82.8\%.

\begin{table}[h]
\centering
\footnotesize
\begin{tabular}{c|cccc}
\toprule
\diagbox[width=8em]{Model}{Training} & Baseline & EMA & Dist$^{*}$ & Dist$^{*}$+EMA \\
\midrule
\net-G & 80.9\% & 82.3\% & 82.2\% & 82.8\% \\
\bottomrule
\end{tabular}%
\caption{Performance improvement with EMA and distillation. *: Distillation-based training}
\label{tab:distill_ema} 
\end{table}

\section{Conclusion}

True to their name, previous neural architecture search methods search only over architectures, using a fixed set of training hyperparameters (i.e., ``training recipe''). As a result, previous methods overlook higher-accuracy architecture-recipe combinations. However, our \nas~does not, being the first algorithm to jointly search over both architectures and training recipes simultaneously for a large dataset like ImageNet. Critically, \nas's~\predictor~pretrains on ``free'' architecture statistics--namely, FLOPs and \#Parameters--to improve the predictor's sample efficiency significantly. After training and using the predictor, the resulting \net~architecture-recipe pairs attain state-of-the-art per-FLOP accuracies on ImageNet classification. 
\footnote{\textbf{Acknowledgments}: Alvin Wan is supported by the National Science Foundation Graduate Research Fellowship under Grant No. DGE 1752814. In addition to NSF CISE Expeditions Award CCF-1730628, UC Berkeley research is supported by gifts from Alibaba, Amazon Web Services, Ant Financial, CapitalOne, Ericsson, Facebook, Futurewei, Google, Intel, Microsoft, Nvidia, Scotiabank, Splunk and VMware.}


{\small
\bibliographystyle{ieee_fullname}
\bibliography{egbib}
}

\clearpage
\appendix

\section{Appendix}

\subsection{Training recipe used in Table~\ref{tab:model_recipe}}
\label{sec:train-recipe-details}
Both Recipe-1 and Recipe-2 share the same batch size of 256, initial learning rate 0.1, weight decay at $4\times 10^{-5}$, SGD optimizer, and cosine learning rate schedule. Recipe-1 train the model for 30 epochs and Recipe-2 train the model for 90 epochs.  We don't introduce training techniques such as dropout, stochastic depth, and mixup in Recipe-1 or Recipe-2.

We make the same observation when training Recipe-1 and Recipe-2 use the same \#Epochs but different weight decay: The accuracy of ResNet18 (1.4x width) is 0.25\% higher and 0.36\% lower than that of ResNet18 (2x depth) when the weight decay is 1$e^{-4}$ and 1$e^{-5}$, respectively.

\subsection{Base architecture in recipe-only search}\label{sec:fbnetv2_l3}
We show the base architecture  (a scaled version of FBNetV2-L2) used in the recipe-only search in Table~\ref{tab:fbnetv2_l3}, while the input resolution is 256$\times$256. This is the base architecture used in the training recipe search in Section~\ref{sec:ros}.  It achieves 79.1\% top-1 accuracy on ImageNet with the original training recipe used for FBNetV2.  With the searched training recipes, it achieves 79.9\% ImageNet top-1 accuracy.

\subsection{Search settings and details}
\label{sec:detailed-settings}
In the \autotrain~search experiment, we set the early-stop rank correlation threshold to be 0.92, and find the corresponding early-stop epoch to be 103.  In the \pbes, we set the population of the initial generation to be 100 (50 best-performing candidates from \cio and 50 randomly generated samples).  We generate 24 children from each candidate and pick the top 40 candidates for the next generation.  Most of the settings are shared by the joint search of architecture and training recipes, except the early-stop epoch to be 108.  The accuracy \predictor~consists of one embedding layer (architecture encoder layer) and one extra hidden layer.  The embedding width is 24 for the joint search (note that there is no pretrained embedding layer for the \autotrain~search). We set both minimum and maximum FLOPs constraint at 400M and 800M for the joint search, respectively.  The selection of $m$ best-performing samples in the constrained iterative optimization involves two steps: (1) equally divide the FLOP range into $m$ bins and (2) pick the sample with the highest predicted score within each bin.

We show the detailed searched training recipe in Table~\ref{tab:searched_recipe}.  We also release the searched models.

\begin{table}[t]
    \footnotesize
    \centering
    \begin{tabular}{l|cc}
    \toprule
    Notation & Value \\
    \hline
    lr & 0.026 \\
    optim & RMSprop \\
    ema & true \\
    p & 0.17 \\
    d & 0.09 \\
    m & 0.19 \\
    wd & 7e-6 \\
    \bottomrule
    \end{tabular}
    \caption{Searched training recipe.}
    \label{tab:searched_recipe}
\end{table}

\subsection{Comparison between recipe-only search and hyperparameter optimizers} Many well-known hyperparameter optimizers (ASHA, Hyberband, PBT) evaluate on CIFAR10. One exception is \cite{metz2020using}, which reports a 0.5\% gain for ResNet50 on ImageNet by searching optimizers, learning rate, weight decay, and momentum. By contrast, our recipe-only search with the same space (without EMA) increases ResNet50 accuracy by 1.9\%, from 76.1\% to 78.0\%.

\subsection{Training settings and details}\label{sec:training_details}
We use distributed training with 8 nodes for the final models,  and scale up the learning rate by the number of distributed nodes (e.g.,  8$\times$ for 8-node training).  The batch size is set to be 256 per node. We use label smoothing and AutoAugment in the training. Additionally, we set the weight decay and momentum for batch normalization parameters to be zero and 0.9, respectively

\begin{figure}[t]
    \centering
    \includegraphics[width=50mm]{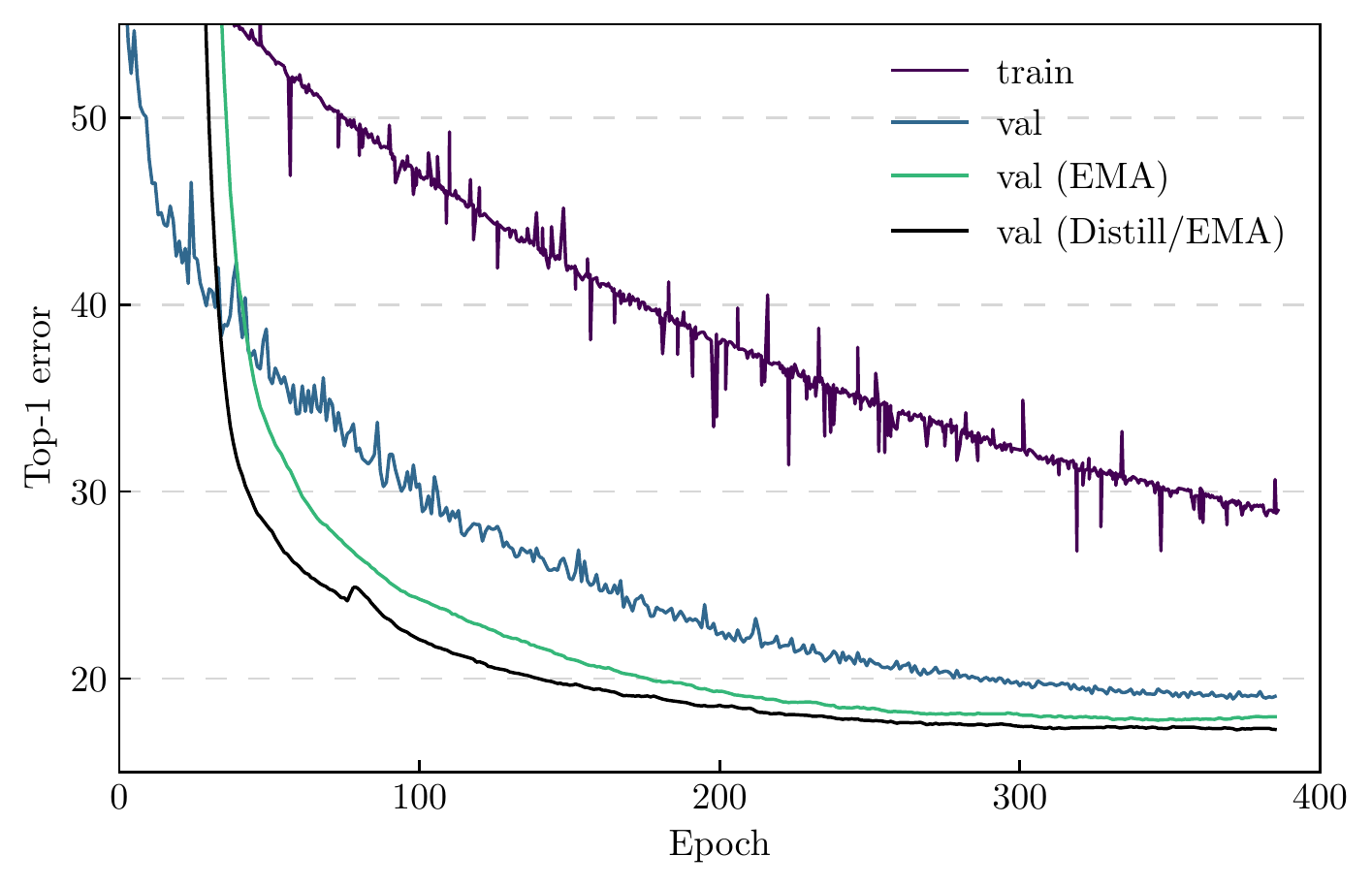}
    \caption{Training curve of the search recipe on \net-G.}
    \label{fig:ema}
\end{figure}

We implement the EMA model as a copy of the original network (they share the same weights at $t=0$). After each backward pass and model weights update, we update the EMA weights as
\begin{equation}
w_{t+1}^{ema} = \alpha w_{t}^{ema} + (1-\alpha)w_{t+1}
\end{equation}
where $w_{t+1}^{ema}$, $w_{t}^{ema}$, and $w_{t+1}$ refer to the EMA weight at step $t+1$, EMA weight at step $t$, and model weight at $t+1$.  We use an EMA decay $\alpha$ of 0.99985, 0.999, and 0.9998 in our experiments on ImageNet, CIFAR-10, and COCO, respectively.  We further provide the training curves of \net-G in Fig.~\ref{fig:ema}.

The baseline models (e.g., AlexNet, ResNet, DenseNet, and ResNeXt) are adopted from PyTorch open-source implementation without any architecture change.  The input resolution is 224$\times$224.

\begin{table*}[t]
\footnotesize
\centering
\resizebox{138mm}{!}{
\begin{tabular*}{145mm}{l @{\extracolsep{\fill}} lllllll}
\toprule
block       & k     &   e   & c     & n     & s  & se & act.\\ \hline
Conv    & 1     & 3     & 16    & 1     & 2  & -  & hswish\\
MBConv 	 & 3 	 & 1                & 16 & 2 & 1 & N & hswish\\
MBConv	 & 5 	 & 5.46  & 24 & 1 & 2 & N & hswish\\
MBConv	 & 5 	 & 1.79  & 24 & 1 & 1 & N & hswish\\
MBConv	 & 3 	 & 1.79  & 24 & 1 & 1 & N & hswish\\
MBConv	 & 5 	 & 1.79  & 24 & 2 & 1 & N & hswish\\
MBConv	 & 5 	 & 5.35  & 40 & 1 & 2 & Y & hswish\\
MBConv	 & 5 	 & 3.54  & 32 & 1 & 1 & Y & hswish\\
MBConv	 & 5 	 & 4.54  & 32 & 3 & 1 & Y & hswish\\
MBConv	 & 5 	 & 5.71  & 72 & 1 & 2 & N & hswish\\
MBConv	 & 3 	 & 2.12  & 72 & 1 & 1 & N & hswish\\
Skip    & - & - & 72 & - & - & - & hswish \\
MBConv	 & 3 	 & 3.12  & 72 & 1 & 1 & N & hswish\\
MBConv	 & 3 	 & 5.03  & 128 & 1 & 1 & N & hswish\\
MBConv	 & 5 	 & 2.51  & 128 & 1 & 1 & Y & hswish\\
MBConv	 & 5 	 & 1.77  & 128 & 1 & 1 & Y & hswish\\
MBConv	 & 5 	 & 2.77  & 128 & 1 & 1 & Y & hswish\\
MBConv	 & 5 	 & 3.77  & 128 & 4 & 1 & Y & hswish\\
MBConv	 & 3 	 & 5.57  & 208 & 1 & 2 & Y & hswish\\
MBConv	 & 5 	 & 2.84  & 208 & 2 & 1 & Y & hswish\\
MBConv	 & 5 	 & 4.88  & 208 & 3 & 1 & Y & hswish\\
Skip    & - & - & 248 & - & - & - & hswish \\
MBPool	 & - 	 & 6	 & 1984 	 & 1                       &- & - & hswish\\
FC     & -    & -     & 1000         & 1        & -           & - & -  \\
 \bottomrule
\end{tabular*}
}
\caption{\footnotesize{Baseline architecture used in the recipe-only search.  The block notations are identical to Table~\ref{tab:search_space}.  Skip block refers to an identity connection if the input and output channel are equal otherwise a 1$\times$1 conv.}}
\label{tab:fbnetv2_l3}
\end{table*}

\subsection{More discussions on training tricks}
\label{sec:training-tricks-efficientnet}

We acknowledge EfficientNet does not use distillation. For fair comparison, we report FBNetV3 accuracy without distillation. We provide an example in Table~\ref{tab:distillation}: Without distillation, FBNetV3 achieves higher accuracy with 27\% less FLOPs, compared to EfficientNet. However, all our training tricks (including EMA and distillation) are used in the other baselines, including BigNAS and OnceForAll.

\begin{table}[t]
\centering
\footnotesize
\begin{tabular}{lcccc}
\toprule
Model & Distillation & FLOPs & Acc. (\%) & $\Delta$\\
\hline
EfficientNetB2 & N & 1050 & 80.3  & 0.0 \\
FBNetV3-E & N & 762 & 80.4 & +0.1\\
FBNetV3-E & Y & 762 & 81.3 & +1.0 \\
\bottomrule
\end{tabular}%
\caption{\small
Model comparison w/ and w/o distillation.}
\label{tab:distillation} 
\end{table}

\begin{figure}[t]
\includegraphics[width=0.45\textwidth]{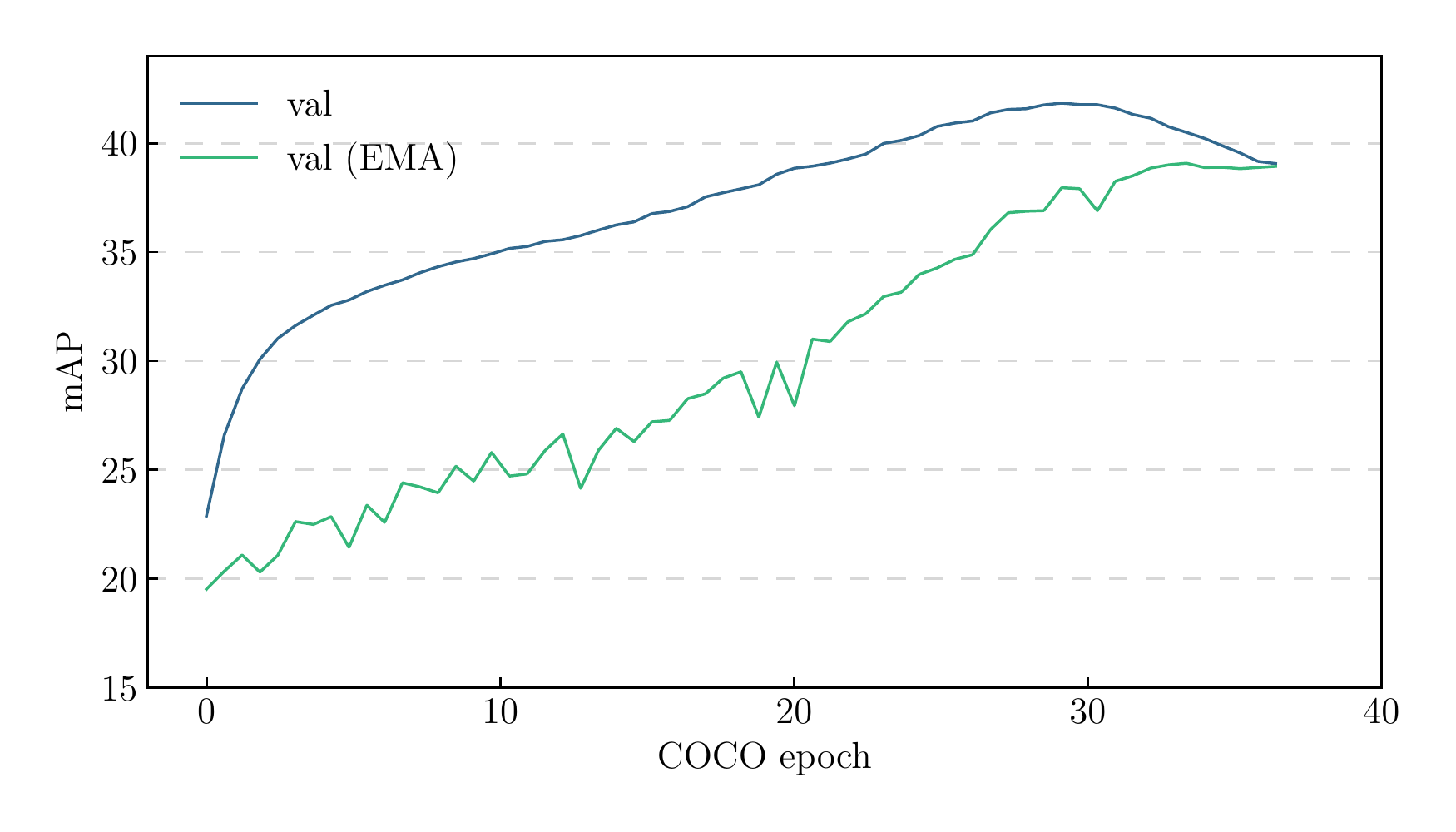}
    \caption{Training curves for RetinaNet with ResNet101 backbone on COCO object detection.}
    \label{fig:training_curve_detection}
\end{figure}

\textbf{Generality of \textit{stochastic weight averaging via EMA}}.  We observe that \textit{stochastic weight averaging via EMA} yields significant accuracy gain for the classification tasks, as has been noted prior \cite{brock2018large,he2020momentum}.  We hypothesize that such a mechanism could be used as a general technique to improve other DNN models. To validate this, we employ it to train a RetinaNet~\cite{retina} on COCO object detection~\cite{coco} with ResNet50 and ResNet101 backbones. We follow most of the default training settings but introduce EMA and Cosine learning rate.  We observe similar training curves and behavior as the classification tasks, as shown in Fig.~\ref{fig:training_curve_detection}.  The generated RetinaNets with ResNet50 and ResNet101 backbones achieve 40.3 and 41.9 mAP, respectively, both substantially outperform the best reported values in~\cite{detectron2} (38.7 and 40.4 for ResNet50 and ResNet101, respectively). A promising future direction is to study such techniques and extend it to other DNNs and applications.

\subsection{Further improvements on FBNetV3}\label{sec:further_gain}
\begin{figure}[h]
    \centering
    \includegraphics[width=83mm]{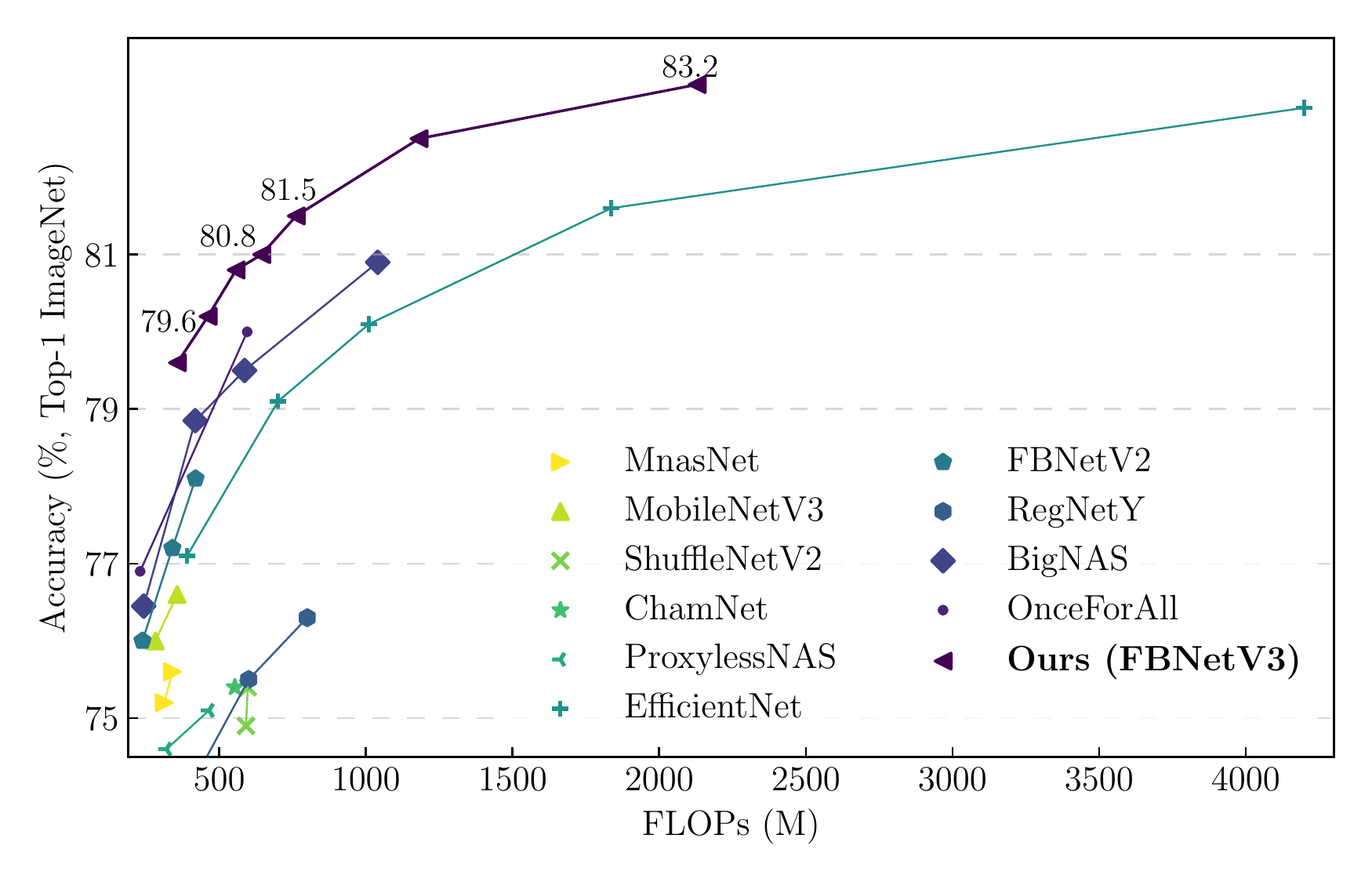}
    \caption{ImageNet accuracy vs. model FLOPs comparison of \net~(distilled from giant RegNet-Y models) with other efficient convolutional neural networks.}
    \label{fig:imagenet_res_regnet}
\end{figure}

We demonstrate that using a teacher model with higher accuracy leads to further accuracy gain on FBNetV3.  We use RegNetY-32G FLOPs (top-1 accuracy 84.5\%)~\cite{giant_regnet} as the teacher model, and distill all the FBNetV3 models.  We show all the derived models in Fig.~\ref{fig:imagenet_res_regnet}, where we observe a consistent accuracy gain at 0.2\% - 0.5\% for all the models.  

\end{document}